\documentclass[sn-mathphys-ay]{sn-jnl}

\usepackage{graphicx}%
\usepackage{multirow}%
\usepackage{amsmath,amssymb,amsfonts}%
\usepackage{amsthm}%
\usepackage{mathrsfs}%
\usepackage[title]{appendix}%
\usepackage{xcolor}%
\usepackage{textcomp}%
\usepackage{manyfoot}%
\usepackage{booktabs}%
\usepackage{algorithm}%
\usepackage{algorithmicx}%
\usepackage{algpseudocode}%
\usepackage{listings}%
\usepackage{hyperref}%

\theoremstyle{thmstyleone}%
\theoremstyle{thmstyletwo}%

\theoremstyle{thmstylethree}%

\begin{document}%

\title[Article Title]{Associativity–Peakiness Metric for Contingency Tables}


\author*[1]{\fnm{Naomi E.} \sur{Zirkind}}\email{naomi.e.zirkind.civ@army.mil}

\author[1]{\fnm{William J.} \sur{Diehl}}\email{william.j.diehl.civ@army.mil}


\affil*[1]{\orgdiv{Army Research Directorate}, \orgname{DEVCOM Army Research Laboratory}, \orgaddress{\street{2800 Powder Mill Road}, \city{Adelphi}, \postcode{20783}, \state{MD}, \country{USA}}}




\abstract{For the use case of comparing the performance of clustering algorithms whose output is a
contingency table, a single performance metric for contingency tables is needed. Such a metric is 
vital for comparative performance analysis of clustering algorithms.  A survey of publicly available 
literature did not show the presence of such a metric. Metrics do exist for vector pairs of truth 
values and predicted values, which are an alternative form of output of clustering algorithms. 
However, the metrics for vector pairs do not reveal the presence of detailed performance 
features that are apparent in contingency tables. This paper presents the Associativity–Peakiness (AP)
 metric, which characterizes aspects of clustering algorithm performance that are critical for predicting 
a clustering algorithm’s performance when deployed. The AP metric is analogous to measures of quality
 for confusion matrices that are outputs of supervised learning algorithms. This paper presents
results from simulations in which 500 contingency tables were generated for multiple test
scenarios. The results show that for the use case of evaluating clustering algorithms, the
AP metric characterizes performance of contingency tables with higher dynamic range than
publicly available metrics, and that it is computationally more efficient than comparable
publicly available metrics.}




\maketitle

\section{Introduction}\label{sec1}

This work addresses the need for a new metric for evaluating the performance of unsupervised
clustering algorithms. Deficiencies of currently used metrics for this purpose are
presented, and a novel metric is presented: associativity combined with peakiness, the AP
metric.

Contingency tables and truth vector–prediction vector pairs are equivalent (assuming element
ordering in the vector pairs does not matter). Demonstration of this fact is presented
in the appendix of \cite{APReport2023}. This equivalence is analogous to the duality    
between time domain and frequency domain representations of a signal. Although a signal can be represented in either domain, each
domain requires its own metrics. For example, bandwidth applies to a frequency-domain
representation. If, for a particular use case, the time domain was most relevant, then time
domain metrics, such as rise time for a pulse, must be defined. In such a case, transforming 
a time-domain function into the frequency domain and then using frequency-domain
metrics on that function would give minimal insight into the rise-time performance of the
function. The metrics presented here are designed specifically for contingency tables. When
the contingency table representation of the data is most relevant for one’s use case, then
metrics for the contingency table must be defined.

The AP metric is quite similar to the figures of merit for a confusion matrix, which is a
primary performance metric for supervised learning algorithms. Desirable characteristics of
confusion matrices are that the diagonal elements are quite large, and that all off-diagonal
elements are quite small. Contingency tables will not, in general, be diagonal for two
reasons: (1) they are not necessarily square matrices, and (2) their two axes are not equal
as is the case with confusion matrices. Therefore, figures of merit for confusion matrices
cannot be applied as-is to contingency tables. The corresponding desirable characteristics
for contingency tables are that they should contain N large numbers, where N is the number
of truth classes, such that each of these N large numbers corresponds to a different cluster.
Also, the other matrix elements besides these N large numbers should be quite small. The
first of these characteristics is measured by the associativity metric, and the second of these
characteristics is measured by the peakiness metric.

The AP metric is critical for conducting theoretical studies of the performance of unsupervised
learning algorithms when different algorithm variants and/or input datasets are
used. As a comparison, in studies of supervised learning algorithms, the performance of
various algorithm variants, possibly with the use of various input datasets, are easily compared
and analyzed by viewing tables with values of scalar metrics such as mean per-class
accuracy, precision, and recall for each algorithm-dataset pair. The AP metric now makes
it possible to perform such studies and performance analysis on unsupervised-learning clustering
algorithms.

The remainder of the paper is organized as follows. The use case of characterizing the
output of clustering algorithms is presented, with a particular contingency table presented
as an example of the need for a metric that quantifies associativity and peakiness. The
next section summarizes the results of a literature search for relevant metrics. It presents
descriptions and equations for metrics from the Python scikit-learn library \cite{bib8} 
and shows the scores that the scikit-learn and AP metrics give to the
particular contingency table that is presented. A detailed description is then presented of
the equations for the AP metric, and of the methodology of evaluating the various metrics
under test. Six test cases are described, and the performance of all the metrics is presented
in each test case. The first two test cases consist of individual contingency tables that
represent ideal performance and worst-case performance. In the remaining four test cases,
score histograms are presented as results of simulations in which 500 contingency tables
were generated for each case. Also, correlation coefficients are presented that show the
correlation of the existing metrics with the AP metric. The correlation coefficients show
to what extent the other metrics under test characterize associativity and peakiness. The
four test cases employ contingency tables of various clustering performance levels. In these
four test cases, various shapes and sizes of contingency tables are used—square as well as
non-square.

\section{Use Case}\label{sec2}

In machine learning, implementation of a supervised learning algorithm is divided into two
main phases: training and inference. During the training phase, the parameters of the
algorithm are optimized so that the algorithm is well fitted to the training data. Once the
algorithm is optimized, the inference phase can commence, in which case a new sample is
input into the algorithm, and the algorithm outputs a prediction regarding the class and/or
value of the input.

Unsupervised learning has analogous phases, namely a research phase and a deployment
phase. During the research phase, the researcher evaluates the performance of various
algorithms, using metrics suited to the particular use case, to see which one performs best
using one or more experimental datasets. Once the best-performing algorithm is identified,
it is deployed to the situation in which it would be used, and it performs its unsupervised
learning tasks, such as clustering, on data that is input into it. This paper addresses metrics
to be used during the research phase of unsupervised learning, as described previously.
Research efforts require performance metrics that would quantify desirable performance
of an algorithm under test, according to the use case for which the algorithm is being
developed.

The use case addressed here is the selection and/or development of an algorithm that
will cluster unrecognized data input samples into clusters composed of similar data samples.
During the research phase, labeled data samples are input into the candidate algorithms,
though the algorithms are not informed of the labels during testing. The labels are retained
only for calculating the performance metrics for each algorithm. Favorable performance of
an algorithm would occur when the data clusters produced from an input dataset closely
resemble the truth clusters that would be formed using the truth labels of each data sample.
An ideal metric for such a research effort is one that quantifies this resemblance.

The contingency table is an informative way to summarize the output of a clustering
algorithm. The axes of a contingency table from unsupervised learning are truth classes and
cluster indices. The contingency table is especially suitable for visualizing and quantifying
the resemblance of the formed clusters with the truth clusters. The metrics presented in
this paper are derived from the entries in a contingency table.

\begin{figure}
\centering
\centerline{\includegraphics[width=0.7\linewidth]{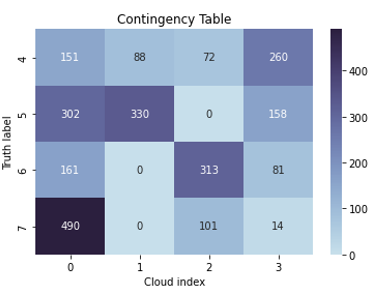}}
\caption{Sample contingency table}
\label{figure1}
\end{figure}

An example of a contingency table that shows the output of a particular clustering
algorithm that was given a particular data set to cluster is shown in Figure~\ref{figure1}.

\begin{figure}
\centering
\centerline{\includegraphics[width=0.7\linewidth]{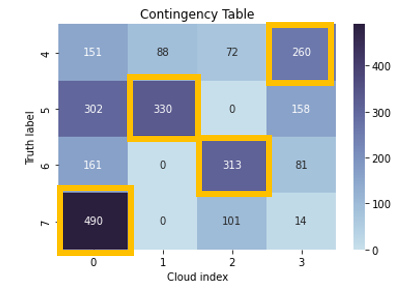}}
\caption{Contingency table highlighting high-associativity squares}
\label{figure2}
\end{figure}

In this paper, the terms ”cluster” and ”cloud” are used interchangeably, and both refer
to the same thing. The table entry squares in Figure~\ref{figure1} are color coded such that the squares
with higher population values are a darker color. With confusion matrices from supervised
learning, one would look for a diagonal matrix. What would one look for in a contingency
table from unsupervised learning? Certainly not a diagonal matrix, since the two axis
labels of the table, “truth label” and “cloud index,” are not equal. A close associativity
between clouds and truth labels would be desirable. Looking at the darkest-colored squares
in Figure~\ref{figure1} shows that truth label 4 is closely associated with cloud 3, truth label 5 is closely
associated with cloud 1, truth label 6 is closely associated with cloud 2, and truth label 7
is closely associated with cloud 0. Figure~\ref{figure2} shows the contingency table of Figure~\ref{figure1} with
each high-associativity square enclosed in a yellow box.

Figure~\ref{figure2} shows a one-to-one associativity between clouds and truth labels. The question
now becomes how to quantify the degree of associativity present in a contingency table.
Furthermore, a way must be found to characterize how distinct the peak in each boxed
value is in Figure~\ref{figure2}. The values of both the associativity and peakiness metrics must range
from 0 to 1.

This paper presents novel metrics for contingency tables that resemble the metrics for
confusion matrices used in supervised learning. In an ideal confusion matrix, all diagonal
elements are large values, and all off-diagonal elements are very small or zero. In a
contingency table, the large-valued elements are not necessarily on the diagonal, but they
should indicate a one-to-one matching as a confusion matrix does. The degree of one-to-one
matching is quantified by the associativity metric. Furthermore, these large-valued metrics
must be large compared with the other elements in the contingency table, as they are in as
confusion matrix. This latter property is quantified by the peakiness metric. Thus, the AP
metric combination captures the fundamental properties of a confusion matrix, which is a
critical performance metric in supervised learning. The next section surveys published literaturess
that is relevant to the goals of finding clustering metrics in the domain of contingency
tables.

\section{Relevant Research Results}\label{sec3}

Much literature has been published regarding associativity for contingency tables, but these works deal with different use cases than the one considered here, which is comparatively evaluating clustering algorithms. This section summarizes some of the less similar use cases as well as a few highly similar use cases.

Two basic papers \cite{bib1}, \cite{bib2} were published by the same author in 1970. One work \cite{bib1} extends previous work on association of rows and columns within a 2x2 contingency table to 2D matrices of arbitrary sizes. The other work \cite{bib2} extends the existing metrics of association to multidimensional contingency tables. 

A comprehensive e-book titled "Contingency Table Analysis" \cite{bib3}  describes a multitude of analysis techniques and metrics for 2x2 tables as well as tables with larger dimensions. These metrics address the correlation or independence between the variables whose data is tabulated in a contingency table. 

Reference~\cite{bib4} evaluates various association coefficients for arbitrarily sized 2D contingency tables. The evaluation framework uses sums of rows and sums of columns, and does not consider individual indices of the table, which could be of significant interest.  Reference~\cite{bib5} gives a comprehensive evaluation of many metrics for 2x2 contingency tables. Such tables measure the correlation between binary variables and do not have relevance for multi-class and multi-cluster situations. 

Reference~\cite{bib6} present a metric for the similarity between attributes (rows) of a contingency table whose entries are categorical values rather than numerical values, which has been thoroughly addressed in prior works. The proposed concept for a similarity measure considers the dependency between attributes listed in the contingency table. In our use case, there is no functional dependency between truth classes or between cluster indices.

In references such as those just cited in this section, the contingency matrix presents results of a survey, such that the rows represent various subgroups of respondents (e.g., males and females), and the columns represent various responses to the survey. It is desired to understand the various correlations—e.g., what percent of those who responded x are males, what percent of females responded y, and so on. These approaches treat the entire contingency table as truth and try to discover correlations that emerge from this true data. In contrast, in our use case, only the class labels are truth, and we want to determine how closely the cluster distribution matches the true distribution.

A paper that is more similar to our use case is  \cite{bib7} which summarizes various pair counting, information theoretic, and entropy-based approaches to comparing two clusterings. Examples of these approaches are the Rand index and mutual information, both of which are part of the Python scikit-learn library \cite{bib8}.  Reference~\cite{bib7} presents a set of desirable characteristics for clustering metrics and evaluates the compliance of various metrics with this set of characteristics. They propose a novel Measure of Concordance metric, which is related to the concepts of precision and recall, and show that it and a few other metrics exhibit the desired behavior in a specific set of test scenarios. However, the only connection of this work to contingency tables is that it uses a 2x2 contingency table to tabulate the results of the pair counting approaches. It does not give any performance metric for an arbitrarily sized contingency table.

Another study, \cite{bib9}, is similar to our use case in that it compares a set of clusters produced by an algorithm with the set of “ground truth” classes. Per that study, their “evaluation scheme quantitatively measures how useful the cluster labels are as predictors of their class labels.”  Reference~\cite{bib9} presents a use case of clustering natural language documents, where the ground-truth labels were assigned by humans. Any algorithm that can produce the same clustering as one made by humans would be highly regarded. This approach uses an entropy-based metric, which measures how useful in bits the clusters are in encoding the class labels. The paper formulates a set of desired characteristics of external metrics and shows that the presented metric satisfies them all in a set of 500 test cases. The only relevance of this work to contingency tables is that, as in  \cite{bib7} a 2x2 contingency table is used to tabulate the results of a binary performance metric. Dom reports, “A value of 0 indicates pairs that were assigned to the same class, whereas a value of 1 corresponds to pairs occurring in different classes.” \cite[p. 138]{bib9}

\subsection{F1 Clustering Metric}  \label{f1 clustering metric}
The most germane work to our use case is in a work on analysis of confusion matrix data from a remote sensing application, \cite{bib10}, which presents two metrics for confusion matrices, which could also be used for contingency tables. Their metrics are called producer's accuracy and user's accuracy. The prodcer's accuracy for each column of the confusion matrix is the maxium value in that column divided by the sum total value of that column. The user's accuracy for each row of the confusion matrix is the maximum value in that row divided by the sum total value of that row. Thus, producer's accuracy corresponds to the truth classes of our contingency tables, and user's accuracy correponds to the clusters of our contingency tables.

The equations used in \cite{bib10}, adapted to the names of the row and columns of contingency tables that show the output of clustering algorithms, are defined by (\ref{eqn1}) and (\ref{eqn2}), respectively.

\begin{equation}
\label{eqn1}
{Truth~Class~Accuracy} = \frac{1}{{N}_{truth~classes}}~~* \\ \\  \sum_{truth~classes}  \frac{max~cluster~population}{sum~of~cluster~ populations}
\end{equation}

\begin{equation}
\label{eqn2}
{Cluster~Purity} = \frac{1}{{N}_{clusters}}~~* \\ \\  \sum_{clusters}  \frac{max~truth~class~population}{sum~of~truth~class~ populations}
\end{equation}

In (\ref{eqn1}), ${N}_{truth~classes}$ is the total number of truth classes. In (\ref{eqn2}), ${N}_{clusters}$ is the total number of clusters. Reference~\cite{bib10} combines the truth class accuracy and the cluster purity metrics into a single metric, called F1, since both metrics are critical in evaluating performance. The F1 metric is the harmonic mean of the two individual metrics.

Although \cite{bib10} presents two metrics for confusion matrices, and combines them into a single metric, the associativity and peakiness metrics presented here have advantages over those metrics, as described later in Section~\ref{compare ap f1}. 

The literature search described in this section shows that there are many studies of metrics of contingency tables in which correlations of two or more variables is quantified, but these results have limited applicability to the use case in which the contingency table represents the results of a clustering algorithm. The works cited in which a contingency table is used in connection with clustering algorithms use only 2x2 contingency tables to represent binary variables that characterize the clustering performance. Only one reference cited \cite{bib10} presents a pair of metrics for contingency tables, but the metrics presented here are more descriptive and comprehensive than those metrics, as described in Section~\ref{compare ap f1}. 

The absence of a single, highly descriptive, publicly available metric for contingency tables is illustrated by a statement in the scikit-learn documentation. This documentation \cite{bib11} lists two drawbacks of contingency tables, the second of which is, “It doesn’t give a single metric to use as an objective for clustering optimisation.” This paper presents and evaluates just such a metric.

\subsection{Scikit-Learn Metrics}
The Python scikit-learn library has a good set of metrics for unsupervised clustering in cases where truth values (also known as “ground truth”) are known \cite{bib8}. These metrics measure the similarity between two clusterings. 

The following scikit-learn metrics are examined in this study for comparison with the novel metrics presented here: Adjusted Mutual Information (AMI), Adjusted Rand Score (ARS), Fowlkes–Mallows Score (FMS), Completeness, Homogeneity, and V-Measure. For each of these metrics, the score it produces is between 0 and 1, except that the ARS ranges from –0.5 to 1 in the scikit-learn implementation \cite{skl3}. Each of these metrics is briefly described in the following paragraphs. Example calculations for all of these metrics are clearly presented in \cite{datasci2023}.

\subsubsection{Adjusted Mutual Information}
The Mutual Information (MI) is a measure of the similarity between two lists of labels of the same data (e.g., predicted [P] and true [T]). MI quantifies the information shared by the two clusterings and therefore can be used as a similarity measure between the two clusterings. A disadvantage of MI is that if the prediction assignments are random, a very undesirable situation, the MI score is nonzero. AMI corrects for chance; its value is 1 when the P and T are identical, and 0 when the MI equals the value due to chance alone. Equations for the metric are in the scikit-learn website \cite{bib8} and in \cite{vinh2009}.

\subsubsection{Adjusted Rand Score}
The Rand Score computes a similarity measure between two clusterings (e.g., predicted [P] and true [T]). The Rand Score is calculated using two variables, a and b, which are defined \cite{skl3} as follows:\\

\begin{list}{}{}
\item {\textit{a} = the number of pairs of elements that are in the same cluster in T and in the same cluster in P} \\
\item{\textit{b} = the number of pairs of elements that are in different clusters in T and in different clusters in P} \\
\end{list}

Using these definitions of $a$ and $b$, the unadjusted Rand Score is given by (\ref{eqn3}), where ${N}_{pairs}$  is the total number of possible pairs in the dataset. 

\begin{equation}
\label{eqn3}
{Rand~Score} = \frac{a+b}{{N}_{pairs}}
\end{equation}

As is the case with MI, the Rand Score gives a nonzero score to random prediction assignments. The ARS \cite{skl3} remedies this deficiency and is defined by (\ref{eqn4}).

\begin{equation}
\label{eqn4}
{ARS} = \frac{RS-E[RS]}{max(RS)- E[RS]}
\end{equation}

\noindent where RS is the Rand Score, and $E[RS]$  is the expected Rand Score of random prediction assignments. As mentioned, the Adjusted Rand Index ranges from –0.5 to 1 in the scikit-learn implementation \cite{skl3}.

\subsubsection{Fowlkes–Mallows Score}
The FMS measures the correctness of a cluster assignment using the geometric mean of the pairwise precision and recall. In particular, the score is derived from three variables that quantify the true positive, false positive, and false negative in a pairwise manner as follows \cite{skl4}:

\begin{list}{}{}
\item {\textit{True Positive (TP)} = the number of pairs of points that belong to the same clusters in both the true labels and the predicted labels} \\
\item {\textit{False Positive (FP)} = the number of pairs of points that belong to the same clusters in the true labels but not in the predicted labels} \\
\item {\textit{False Negative (FN)} = the number of pairs of points that belong to the same clusters in the predicted labels but not in the true labels} \\
\end{list}

Using these three definitions, the FM score is defined by (\ref{eqn5}).
\begin{equation}
\label{eqn5}
{FM~Score} = \frac{TP}{\sqrt{(TP +  FP) * (TP +  FN)}}
\end{equation}

\subsubsection{Completeness, Homogeneity, and V-Measure Score} \label{completeness etc}
A clustering result satisfies Completeness if all the data points that are members of a given class are elements of the same cluster. A clustering result satisfies Homogeneity if all of its clusters contain only data points that are members of a single class. The degree of Completeness or Homogeneity for a particular clustering is based on conditional entropy considerations. The equations for the Completeness Score and Homogeneity Score are in \cite{rosenberg2007}.

The V-measure is the weighted harmonic mean between Homogeneity and Completeness \cite{rosenberg2007} as shown in  (\ref{eqn6}).

\begin{equation}
\label{eqn6}
V = \frac{(1 + \beta ) * homogeneity * completeness}{\beta * homogeneity + completeness}
\end{equation}

In (\ref{eqn6}), the $\beta$  parameter can be used to give greater weight to Homogeneity as compared to Completeness. In the scikit-learn implementation, the default value of $\beta$  is 1. In the results shown in this paper, this default value is used.

\section{Application of Current Metrics to Particular Contingency Table}
The six clustering metrics described previously were all applied to the contingency table of Figure \ref{figure1} in order to quantify the clustering effectiveness of the algorithm that produced the data in that contingency table. The results are shown in Table~\ref{table1}.

\begin{table} [h]
\caption{scikit-learn scores for contingency table in Figure~\ref{figure1}}\label{table1}
\label{table1}
\begin{tabular}{@{}ll@{}}
\toprule
Metric &  Score\\
\midrule
AMI & 0.246 \\
ARS & 0.162 \\
FMS & 0.370 \\
Completeness & 0.230 \\
Homogeneity & 0.267 \\
V-Measure & 0.275 \\
\botrule
\end{tabular}
\end{table}

The scikit-learn metrics all gave low scores to this contingency table, ranging from 0.162 to 0.370 on a scale of 0 to 1. This finding shows that the scikit-learn metrics do not characterize the associativity between the clusters formed and the truth clusters, at least for this contingency table. This deficiency in the scikit-learn metrics prompted development of new metrics that would quantify this associativity between cluster number and truth class. The next section describes the procedure for calculating the novel metrics, associativity and peakiness.

\section{Associativity–Peakiness Metric}
The novel metrics are called the associativity metric and the peakiness metric. The associativity metric uses data from the contingency table to quantify the degree of matching between the clusters formed by the algorithm and the truth classes. The contingency table in Figure~\ref{figure1} would get an associativity score of 1. In contrast, a contingency table in which all data samples are in a single cluster would get an associativity score of 0. The associativity metric quantifies the associativity in all intermediate cases as well. 

The associativity metric selects, for each truth class, the cluster with the highest population, as shown in Figure~\ref{figure2}, in order to calculate an associativity score. An additional metric is required to show the significance of that highest population value. If that highest value was just slightly above the populations of the other clusters for that truth class, then that highest value does not have much significance, and the associativity metric score derived from that highest value also does not have much significance. On the other hand, if the population values other than that highest value were all zero, then that highest value has very high significance. The metric used to quantify the significance of the highest cluster population for each truth class is called the peakiness metric—it shows how peaky that highest population value is.

\subsection{Associativity Metric}
This section shows the equations for calculating the associativity metric.

Let the contingency table be denoted at a matrix ${T}_{i,j}$ where the rows $i$ are the truth classes and the columns $j$ are the cluster indices. The first step is to find, for each truth class, the cluster number with the largest value, and to form an unordered list $L$ of these numbers as defined in (\ref{eqn7}). 

\begin{equation}
\label{eqn7}
L = \{ {argmax}_{j}({T}_{i,j}) \}
\end{equation}

The list $L$ in (\ref{eqn7}) is then used to form an unordered list $M$ of all possible pairs of distinct elements of list $L$ , as defined in (\ref{eqn8}).

\begin{equation}
\label{eqn8}
M = \{ ({L}_{m}, {L}_{n}), m \neq n \}
\end{equation}

The associativity metric $A$ is derived from the list $M$ as shown in (\ref{eqn9}). 

\begin{equation}
\label{eqn9}
A =\frac{Number~of~pairs~in~M~whose~members~are~unequal}{Total~number~of pairs~in~M}
\end{equation}

\begin{figure} 
 \centerline{\includegraphics[width=0.7\linewidth]{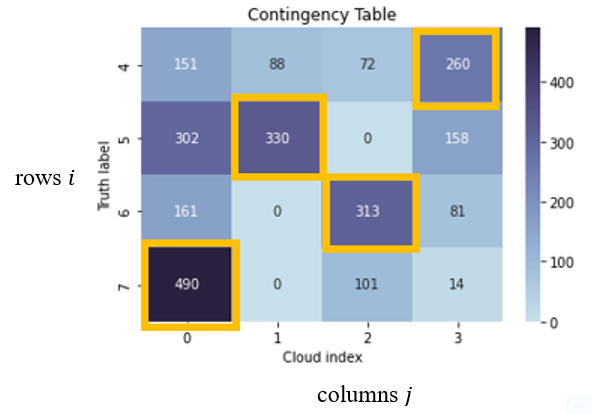}}
  \caption{Sample contingency table with rows and columns labeled} 
 \label{figure3} 
\end{figure}

An example of how the associativity metric is computed for the contingency table in Figure~\ref{figure1} is shown in Figure~\ref{figure3}.

For each truth class, find the cluster number with the largest value, resulting in $ L= [3,1,2,0]$.

Form the list of all possible distinct pairs of elements of $L$, resulting in \\$M= [(3,1),(3,2),(3,0),(1,2),(1,0),(2,0)]$.

The number of pairs in $M$ whose members are unequal is 6. Total number of pairs in $M$ is 6. Therefore, the associativity metric $A=  6/6=1$.

\subsection{Peakiness Metric}
This section shows the equations for calculating the peakiness metric. This metric tells how much of an outlier are the elements of the list of $L$. For example, in the contingency table in Figure~\ref{figure3}, in the second row of this table, the maximum value of 330 is not much larger than one of the other values, 302, in the same row. Thus, the value of 330 is not much of an outlier in the elements of that row. In contrast, in the bottom row of Figure~\ref{figure3}, the maximum value of 490 is much larger than all the other values in that row, so it is a significant outlier. The peakiness metric tells how peaky the row maxima are, and this metric gives an indication of the confidence level of the peaks that are used to calculate the associativity metric.

When calculating the peakiness metric, the largest and second largest values are found for each row (corresponding to a truth class) of the contingency table. The notations shown in (\ref{eqn10}) and (\ref{eqn11}) are used to denote the largest value and second largest value in a list $N$, respectively. 

\begin{equation}
\label{eqn10}
{max}_{1}(N) = largest~value~of~list~N
\end{equation}

\begin{equation}
\label{eqn11}
{max}_{2}(N) = second~largest~value~of~list~N
\end{equation}

For each truth value, the two largest values are found, and the peakiness value for the elements in row i is given by (\ref{eqn12}).

\begin{equation}
\label{eqn12}
{P}_{i} = \frac{{max}_{1}({row}_{i}) - {max}_{2}({row}_{i})} {{max}_{1}({row}_{i})}
\end{equation}

Note that the expression in (\ref{eqn12}) is guaranteed to be between 0 and 1. It is 0 when the largest and second largest values in a row are equal. It is less than or equal to 1 by construction, since ${max}_{1}({L}_{i})$  and ${max}_{2}({L}_{i})$ are both nonnegative, and therefore the numerator is by definition less than or equal to the denominator. 

Special care must be taken in the case when ${max}_{1}({row}_{i}) = 0$, which would happen only when all elements of a row are zero. In that case, the peakiness metric is undefined for that row, so that row is excluded from the calculation of the peakiness metric. A possible alternative for this case is to define the peakiness metric for that row as 0, as would be the case for any other row for which the two largest elements are equal.

The overall peakiness metric is the mean of each of the row values given by (\ref{eqn12}) and is shown in (\ref{eqn13}).

\begin{equation}
\label{eqn13}
P = mean(\{ {P}_{i} \} )
\end{equation}

An example of how the peakiness metric is computed for the contingency table in Figure~\ref{figure1} is shown in Table~\ref{table2}. For each truth class, the largest value and the second largest value are listed. The resulting calculation of the peakiness metric for each row is shown in Table~\ref{table2}.

\begin{table}[h]
\caption{Peakiness metric calculation for sample contingency table}
\label{table2}
\begin{tabular}{@{}llll@{}}
\toprule
Truth class & Largest & Second largest & ${P}_{i}$ using\\
number & value & value &  (\ref{eqn12}) \\
\midrule
4 & 260 & 151 & 0.419 \\
5 & 330 & 302 & 0.085 \\
6 & 313 & 161 & 0.486 \\
7 & 490 & 101 & 0.794 \\
\botrule
\end{tabular}
\end{table}

Note the much higher peakiness value for truth class 7 (0.794) than for truth value 5 (0.085). The overall peakiness value is the mean of all the row values ${P}_{i}$  and is 0.446.

This section has presented the equations and sample calculations for the associativity and the peakiness metrics. The associativity metric is critical since it measures the associativity between the clusters formed by an algorithm with the truth classes. The peakiness metric is critical since it measures the confidence in the peak values used to calculate the associativity metric. Because both of these metrics are critical in quantifying the performance of an algorithm as shown in a contingency table, a composite metric is defined that combines both of these metrics into a single metric—the associativity–peakiness (AP) metric.

\subsection{Associativity–Peakiness Metric}
The AP metric is defined as the harmonic mean of the associativity metric (A) and the peakiness metric (P), analogous to the definition of the V-Measure metric as the harmonic mean of the Completeness and the Homogeneity metrics, \cite{rosenberg2007}, and the definition of the F1 metric as the harmonic mean of truth class accuracy and cluster purity as described earlier in Section~\ref{f1 clustering metric}. The AP metric is defined by (\ref{eqn14}).

\begin{equation}
\label{eqn14}
AP = 2 * \frac{A * P} {A + P}
\end{equation}

For the contingency table shown in Figure~\ref{figure1}, for which A = 1 and P = 0.446 as shown in the sample calculations above,  (\ref{eqn14}) gives an AP value of 0.617. 

Now we can compare the AP score of the contingency table of Figure~\ref{figure1} with the scores of the scikit-learn metrics shown in Table~\ref{table1}, as well as with the F1 score, as calculated from (\ref{eqn1}) and (\ref{eqn2}). The combined list of metrics is shown in Table~\ref{table3}.

\begin{table}[h]
\caption{All metrics scores for contingency table in Figure~\ref{figure1}}
\label{table3}
\begin{tabular}{@{}ll@{}}
\toprule
Metric & Score \\
\midrule
AP & 0.617 \\
AMI & 0.246 \\
ARS & 0.162 \\
FMS & 0.370\\
Completeness & 0.230 \\
Homogeneity & 0.267 \\
V-Measure & 0.275 \\
F1 & 0.578 \\
\botrule
\end{tabular}
\end{table}

Table~\ref{table3} shows that for the contingency table of Figure~\ref{figure1}, the AP metric gives a somewhat favorable rating (0.617), whereas all the scikit-learn metrics gave very unfavorable ratings to this contingency table. The F1 metric, \cite{bib10}, which was designed for use with confusion matrices or contingency tables, gave a fairly favorable rating, close to the score that the AP metric gave. Thus, in the case of the contingency table of Figure~\ref{figure1}, metrics designed for use with contingency tables performed better than those designed for use with truth–prediction vector pairs. These results are anecdotal; later sections compare the performance of the AP metric versus the scikit-learn metrics and the F1 metric for a much larger set of contingency tables. 

\section{Methods}
This section presents the research questions addressed in this study to demonstrate the benefits of the AP metric. It explains the rationale for the data presented in order to address these reseearch questions, the methodology for specifying and generating the data, and the results that address these research questions.

When an unsupervised clustering algorithm is deployed, the only result the user obtains is the set of clusters: the number of clusters and the population of each cluster. There are some underlying truth categories to this data, but these categories are unknown to the user. The user can only assume and hope that each cluster can be associated with its unknown truth category with high confidence. The best way to ensure that this association during deployment is valid is to use a metric that measures associativity as a way to select the best-performing algorithm during the research phase. A well-designed associativity metric could support such an assurance. 

The AP metric is a good candidate since it closely emulates the metrics used for confusion matrices that are used with supervised learning. The desirable features of a confusion matrix are that the diagonal elements are relatively large, and the off-diagonal elements are much smaller than the diagonal elements. In a contingency table, the most populous entries are not necessarily along the diagonal—and there might not even be a diagonal since the contingency table could be non-square. However, it is desirable that there be a one-to-one associativity between the most populous entries and the truth classes, just as there is in a confusion matrix that indicates good performance. The associativity metric quantifies this associativity. Regarding the desirable quality of the off-diagonal elements of a confusion matrix being much smaller than the diagonal elements, the peakiness metric quantifies this behavior for contingency tables. Thus, the AP metric is to contingency tables what diagonality is to confusion matrices.

The objective of presenting the data shown below in this section is to answer the following two questions: 

\begin{enumerate}
\item{What does the AP metric tell us that the scikit-learn and F1 metrics do not?}
\item{How well do the scikit-learn metrics and the F1 metric correlate with the AP metric (i.e., how well do they evaluate associativity and peakiness)? }
\end{enumerate}

\subsection{Evaluation Methodology} \label{eval method}
All of the performance metrics considered in this paper (i.e., the AP metric, the six scikit-learn metrics, and the F1 metric) were applied to six test scenarios:

\begin{enumerate}
\item{Extreme Case – Ideal Performance}
\item{Extreme Case – Worst Performance}
\item{Low Performance, 4x4 contingency tables}
\item{Higher Performance, 4x4 contingency tables}
\item{Higher Performance, 4x6 contingency tables}
\item{Higher Performance, 4x2 contingency tables}
\end{enumerate}

In the first two test scenarios, a single contingency table was used. In the ideal performance case, the contingency table represents an ideal clustering result. In the worst-case performance scenario, the contingency table represents a very poor clustering result. These extreme scenarios are used to examine the basic capability of the metrics to properly characterize an extreme case. If a metric cannot correctly characterize one or both scenarios, then it cannot be effectively used for characterizing clustering performance as expressed in a contingency table.

The low performance case uses 500 randomly generated contingency tables that inherently represent low performance since the table entries were randomly generated. In this test case, a 4x4 contingency table was used.

The three higher-performance test scenarios each use 500 contingency tables that were randomly generated with a high weight placed on using zeros for the contingency table entries. The presence of these several zeros, while keeping the total of all entries fixed, forces the nonzero entries to be larger. A higher-performance contingency table has several large-valued entries, so this method of generating contingency tables more closely resembles high-performance tables as compared with tables whose entries are all randomly generated. 

Three sizes of higher-performance test scenarios were used to explore the performance of the metrics with non-square contingency tables. The three table sizes are 4x4, as used in the extreme case scenarios and in the low-performance scenario, as well as 4x6 and 4x2. In each scenario, the number of truth classes is fixed at 4, and the number of clusters varies between 2 and 6.

As mentioned, for each test scenario, 500 contingency tables were randomly generated. The sum total of all elements in each contingency table is 2521, which is the sum total of all elements in the contingency table shown in Fig. 1. The number 2521 represents the total number of test samples that was used in the clustering process. 

Each contingency table in test scenarios 3–6 was created by first generating a vector of random numbers with a specified number of elements and a fixed sum total (i.e., 2521). This vector was then reshaped to a 2D matrix, with dimensions as required by the test scenario. 

For test scenarios 3–6, the following two data products are presented: 

\begin{itemize}
\item{Histograms of score distributions for each metric}
\item{Bar plot of correlation coefficients between each metric and the AP metric}
\end{itemize}

\subsection{Method for Generating the Random Vectors}  \label{data generation}
Here we explain the generation of random vectors, which are reshaped to form contingency tables. The inputs to this process are the sum total of all elements in the vector (total) and the number of elements in each vector (numvals). Two Python statements \cite{dickinson2023},  (\ref{eqn15}) and (\ref{eqn17}), were used to generate a random vector based on these two inputs.

\begin{equation}
\label{eqn15}
Dividers = 
\\sorted(random.choices(range(0, total), k=numvals-1))
\end{equation}

Equation (\ref{eqn15}) uses the Python “random” package \cite{python2023}, which generates a set of pseudo-random numbers within a specified range, given a specified number of values to generate. Equation (\ref{eqn15}) creates an ordered set of dividers, whose form is shown in (\ref{eqn16}).

\begin{equation}
\label{eqn16}
dividers= \{ {d}_{0}, {d}_{1}, {d}_{2}, ... , {d}_{numvals-1} \}
\end{equation}

To generate higher performance contingency tables, the optional argument $weights$ parameter in the $random.choices$ function was used. The $weights$ parameter is a vector of relative weights for each of the elements in the dividers set \cite{python2023}. In this study, the value 0 was given a weight of 1000, and the value of all the rest of the elements in $dividers$ was given a weight of 1.

The second Python statement from \cite{dickinson2023} that was used to create the random vectors is shown in (\ref{eqn17}).

\begin{equation}
\label{eqn17}
arr = [a – b~for~ a, b~ in~zip(dividers + [total], [0] + dividers)]
\end{equation}

The input to the \textbf{zip} function in (\ref{eqn17}) is a pair of lists that are derived from the list $dividers$ from (\ref{eqn16}). Using the notation of (\ref{eqn16}), these two lists are shown in (\ref{eqn18}) and (\ref{eqn19}).

\begin{equation}
\label{eqn18}
List 1 : [{d}_{0}, {d}_{1}, {d}_{2}, ... , {d}_{numvals-1}, total]
\end{equation}
\begin{equation}
\label{eqn19}
List 2 : [0, {d}_{0}, {d}_{1}, {d}_{2}, ... , {d}_{numvals-1}]
\end{equation}

The \textbf{zip} function creates a list of corresponding pairs, one from each list, as shown in (\ref{eqn20}).

\begin{equation}
\label{eqn20}
 [({d}_{0},0), ({d}_{1}, {d}_{0}) ,  ({d}_{2}, {d}_{1}) ,... , (total, {d}_{numvals-1}) ]
\end{equation}

The vector $arr$, which is the output of  (\ref{eqn17}), is a list of differences of the two elements of each list shown in  (\ref{eqn20}), resulting in the expression shown in  (\ref{eqn21}).

\begin{equation}
\label{eqn21}
 arr = [{d}_{0},  {d}_{1}- {d}_{0},  {d}_{2}- {d}_{1} ,... , total - {d}_{numvals-1}]
\end{equation}

An inspection of (\ref{eqn21}) shows that the sum of the elements in $arr$ is indeed $total$. The vector $arr$ is reshaped to form a contingency table with the desired dimensions. When higher-performance contingency tables are formed, the order of the elements in $arr$ is shuffled so that the zero values are distributed throughout the contingency table rather than all in the first one or few rows.

\section{Results}
The subsections in this section present the results for the six test cases enumerated in Section~\ref{eval method}. 

The first two test cases are analyses of the extreme cases of a 4x4 contingency table with ideal performance, and a 4x4 contingency table with worst-case performance. 

Each of the remaining four test cases is an analysis of the distribution of the metrics using data with a particular performance level and particular contingency table dimensions. The four test cases are: low performance 4x4 contingency tables, higher performance 4x4 contingency tables, higher performance 4x6 contingency tables, and higher performance 4x2 contingency tables.

\subsection{Results: Extreme Cases}
\begin{figure} 
 \centerline{\includegraphics[width=0.5\linewidth]{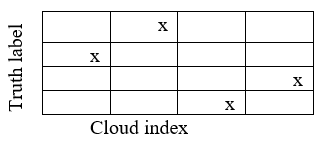}}
  \caption{Structure of a contingency table with ideal performance} 
 \label{figure4} 
\end{figure}

The ideal performance for a confusion matrix from a supervised learning algorithm is a diagonal matrix. The comparable ideal performance for a 4x4 contingency table from a clustering algorithm, in which the rows are truth classes and the columns are cluster indices, is that four of the elements are nonzero, the rest of the elements are zero, and there is a one-to-one matching between the truth classes and the columns. Figure~\ref{figure4} displays the structure of such a contingency table that shows ideal performance. 

In Figure~\ref{figure4}, the x’s represent nonzero values, and the blank entries contain zero values. In this perfect case, the associativity metric is 1, as can be derived from (\ref{eqn7}), (\ref{eqn8}), and (\ref{eqn9}). The associativity metric characterizes the placement of the row maxima.

\begin{figure}
\centering
\includegraphics[width=0.5\linewidth]{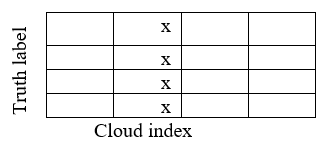}
\caption{Structure of a contingency table with worst-case performance}
\label{figure5}
\end{figure}

In contrast to ideal performance, the worst-case performance occurs when all of the nonzero entries fall into one particular cluster, as shown in Figure~\ref{figure5}.

In such a case, the clustering algorithm gives no information regarding the association of the data samples with the various truth classes. The associativity metric gives this situation a zero score. Since the associativity metric is zero, by construction (harmonic mean, (\ref{eqn14})), the AP metric is also zero. 

\begin{figure}
\centering
\includegraphics[width=0.5\linewidth]{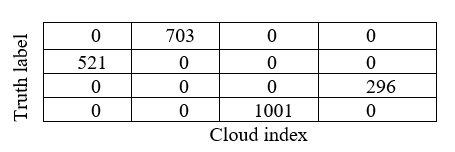}
\caption{Contingency matrix 1, ideal performance}
\label{figure6}
\end{figure}

\begin{figure}
\centering
\includegraphics[width=0.5\linewidth]{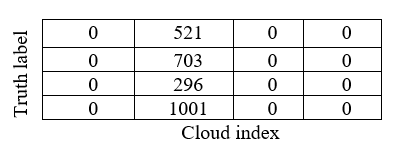}
\caption{Contingency matrix 2, worst-case performance}
\label{figure7}
\end{figure}

Now that ideal and worst-case performance has been defined, sample contingency tables with the structures defined in Figure~\ref{figure4} and Figure~\ref{figure5} can be defined and evaluated by all of the metrics. Figure~\ref{figure6} and Figure~\ref{figure7} show sample contingency tables that emulate the contingency table structures shown in Figure~\ref{figure4} and Figure~\ref{figure5}, respectively, and that show ideal performance and worst-case performance, respectively. The sum total of all the elements in both of these metrics is 2521, which is the same sum total as the contingency table in Figure~\ref{figure1}.

The scores that each of the metrics gave to the two contingency tables of  Figure~\ref{figure6} and Figure~\ref{figure7}  are shown in Table~\ref{table4}.

\begin{table}[h]
\caption{Metrics scores for ideal and worst-case contingency tables}
\label{table4}
\begin{tabular}{@{}lll@{}}
\toprule
Metric & Matrix~1 & Matrix~2~score \\
 & score~(ideal) & (worst~case) \\
\midrule
AP & 1.0 & 0.0 \\
AMI & 1.0 & 0.0 \\
ARS & 1.0 & 0.0  \\
Completeness & 1.0 & \textbf{1.0} \\
Homogeneity & 1.0 & 0.0 \\
V-Measure & 1.0 & 0.0 \\
FMS & 1.0 & \textbf{0.540} \\
F1 & 1.0 & \textbf{0.568} \\
\botrule
\end{tabular}
\end{table}

Table~\ref{table4} shows that all of the metrics correctly scored the ideal case. Almost all of the scikit-learn metrics correctly scored the worst case. The AMI score was not exactly zero, at 7.0 E-16, but was very close to it. The Completeness metric gave a 1.0 score to the worst-case metric, as it should according to its definition (Section~\ref{completeness etc}). However, the V-Measure metric, which combines Completeness and Homogeneity, does give a zero score. The FMS metric gave a very poor score, 0.540. The F1 metric, which is not a scikit-learn metric, and which was designed for use with contingency tables, gave an even poorer score at 0.568. The nonzero scores for the worst-case contingency table are in bold font in Table~\ref{table4} to highlight their presence. 

Inspecting the equations for the FMS ((\ref{eqn5})) and the F1 metrics ((\ref{eqn1}) and (\ref{eqn2})) shows that their definitions would not give a zero score to the worst-case contingency table of Figure~\ref{figure7}. In general, the FMS would give a score of zero  when no two data samples belong to the same cluster in the truth listing and/or in the predicted listing. For example, when every single data sample is erroneously classified, then the FMS would be zero. The F1 metric would be zero only for a contingency table whose maximum cluster population is zero for all truth classes, and/or whose maximum truth class population is zero for all clusters. Both of these cases imply a contingency table, all of whose values are zero.

\subsection{Results: Low-Performance Case}
\begin{figure}
\centering
\includegraphics[width=0.8\linewidth]{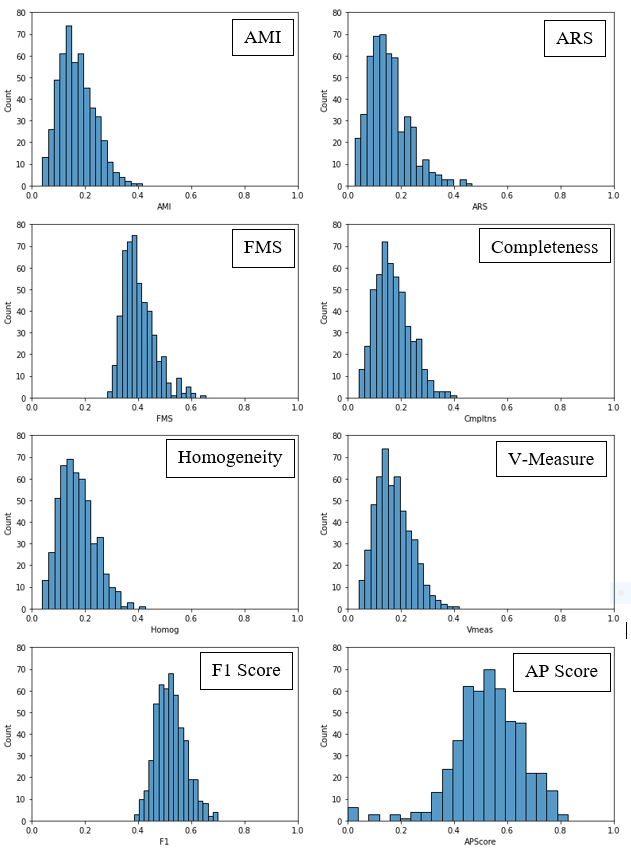}
\caption{Histogram distributions for metrics: low performance, 4x4}
\label{figure8}
\end{figure}

Figure~\ref{figure8} shows, for each of the metrics under test, the histogram of scores for the 500 randomly generated contingency tables. The metrics whose plots are shown in Figure 8 are AMI, ARS, FMS, Completeness, Homogeneity, V-Measure, F1 Score, and the AP Score. The scores that were used to construct the histograms were derived from 4x4 contingency tables with low performance, as described in Section~\ref{eval method}.

The histograms In Figure~\ref{figure8} show that ARS and the entropy-based metrics, AMI, Completeness, Homogeneity, and V-Measure, gave low scores to the set of contingency tables—the most common value was about 0.15 for all of them. However, none of these metrics gave a zero score to any of the contingency tables.

On the other hand, the metrics not based on entropy, FMS and F1, gave much higher scores—their peak values were 0.4 and 0.5, respectively. Furthermore, the range of values of FMS and F1 was only about 0.3 wide, slightly narrower than the distributions from ARS and the entropy-based metrics. FMS values ranged from 0.3 to 0.6, and F1 values ranged from 0.4 to 0.7. 

Only the AP metric gave some zero scores, but it also gave many higher scores such that its peak value was about 0.55 and its maximum value was greater than 0.8. None of the metrics besides AP gave such a wide range of scores. The AP metric shows a higher dynamic range, and thus a higher sensitivity, than the other metrics.

To better understand the behavior of the various metrics, especially at the extremes of high scores and zero scores, four specific contingency tables of the 500 generated samples from this test case are examined in more detail in the next section. 

\subsubsection{Analysis of Selected Contingency Tables}
In this study, 500 contingency tables were generated for each test scenario. For the low-performance scenario, in which the contingency tables contained few zeros, four contingency tables were selected for inspection of their values, and for 
examination of scores assigned to them by the various metrics. Two of these selected contingency tables had the two highest AP scores, and two of them had AP scores of zero. 

For each of the contingency tables presented, the rows are truth values, and the columns are cluster indices. For each row (i.e., truth value) in these contingency tables, the maximum value is highlighted to facilitate understanding of the scores of the metrics.

\begin{figure}
\centering
\includegraphics[width=2.5in]{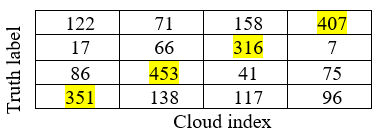}
\caption{Sample contingency table 1, highest AP score}
\label{figure9}
\end{figure}

\textbf{Highest AP score.} Figure~\ref{figure9} shows the contingency table with the highest AP score of the entire set of contingency tables that were generated. Table~\ref{table5} shows the scores that each of the metrics gave to the contingency table shown in Figure~\ref{figure9}.

\begin{table}[h]
\caption{Metrics scores for sample contingency table 1}
\label{table5}
\begin{tabular}{@{}ll@{}}
\toprule
Metric & Score \\
\midrule
AP & 0.827  \\
AMI & 0.237 \\
ARS & 0.231  \\
FMS & 0.428 \\
Completeness & 0.237 \\
Homogeneity & 0.240\\
V-Measure & 0.238 \\
F1 & 0.617 \\
\botrule
\end{tabular}
\end{table}

In Figure~\ref{figure9}, there is a one-to-one matching between clusters and truth classes, resulting in an associativity score of 1. For each row of the contingency table in Figure~\ref{figure9}, the peak is much larger than all the other values in that row, resulting in a high peakiness score, and thus a high AP score of 0.827. 

The entropy-based metrics and ARS all gave scores of between 0.2 and 0.3 to this contingency table, except for FMS and F1, which tend to give higher scores for this dataset. However, their scores, 0.428 and 0.617, do not come near the score that the AP metric gave to this very high-performing contingency table.

\begin{figure}
\centering
\includegraphics[width=2.5in]{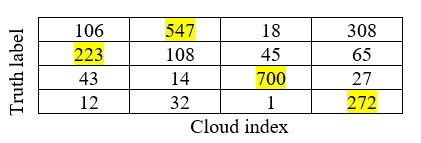}
\caption{Sample contingency table 2, second-highest AP score}
\label{figure10}
\end{figure}

\textbf{Second-highest AP score.} Figure~\ref{figure10} shows the contingency table that generated the second-highest AP score of the entire set of contingency tables that were generated. Table~\ref{table6} shows the scores that each of the metrics gave to the contingency table shown in Figure~\ref{figure10}.

\begin{table}[h]
\caption{Metrics scores for sample contingency table 2}
\label{table6}
\begin{tabular}{@{}lll@{}}
\toprule
Metric & Score \\
\midrule
AP & 0.819  \\
AMI & 0.417 \\
ARS & 0.429  \\
FMS & 0.588 \\
Completeness & 0.408 \\
Homogeneity & 0.427\\
V-Measure & 0.417 \\
F1 & 0.694 \\
\botrule
\end{tabular}
\end{table}

The contingency table in Figure~\ref{figure10} shares the good qualities of the contingency table in Figure~\ref{figure9}, though the peakiness scores of the first and second rows in Figure~\ref{figure10} are not so high, resulting in a slightly lower AP score of 0.819 as compared to the contingency table in Figure~\ref{figure9}. In contrast to Table~\ref{table5} for the contingency table of Figure~\ref{figure9}, Table~\ref{table6} shows that the entropy-based metrics and ARS gave higher scores (between 0.4 and 0.5) to the Figure~\ref{figure10} contingency table than they did to the Figure~\ref{figure9} contingency table. The F1 metric also gave a slightly higher score (0.694 vs. 0.617) to the Figure~\ref{figure10} contingency table than it did to the contingency table in Figure~\ref{figure9}. 

\begin{figure}[!t]
\centering
\includegraphics[width=2.5in]{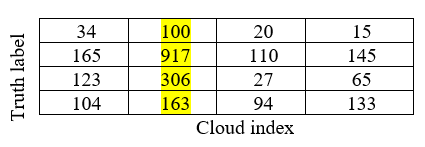}
\caption{Sample contingency table 3, zero AP score}
\label{figure11}
\end{figure}

\textbf{AP score of zero.} Figure~\ref{figure11} shows a contingency table that gave an AP score of zero. Table~\ref{table7} shows the scores that each of the metrics gave to the contingency table shown in Figure~\ref{figure11}.

\begin{table}[h]
\caption{Metrics scores for sample contingency table 3}
\label{table7}
\begin{tabular}{@{}ll@{}}
\toprule
Metric & Score \\
\midrule
AP & 0.0  \\
AMI & 0.040 \\
ARS & 0.095  \\
FMS & 0.444 \\
Completeness & 0.042 \\
Homogeneity&  0.040\\
V-Measure & 0.041\\
F1 & 0.502 \\
\botrule
\end{tabular}
\end{table}

Almost all of the metrics gave very low scores—less that 0.1—to the contingency table of Figure~\ref{figure11}, though only the AP metrics gave it a score of zero. The only two metrics that gave higher scores to this contingency table are FMS at 0.444, and F1 at 0.502. These two metrics do not regard the occurrence of all truth-class peaks in the same cluster, as shown in Figure~\ref{figure11}, as an undesirable phenomenon. 

\begin{figure}
\centering
\includegraphics[width=2.5in]{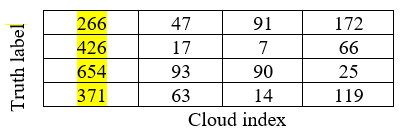}
\caption{Sample contingency table 4, zero AP score}
\label{figure12}
\end{figure}

\textbf{Another case with AP score of zero.}  Figure~\ref{figure12} shows a second contingency table that gave an AP score of zero. Table~\ref{table8} shows the scores that each of the metrics gave to the contingency table shown in Figure~\ref{figure12}.

\begin{table}[h]
\caption{Metrics scores for sample contingency table 4}
\label{table8}
\begin{tabular}{@{}ll@{}}
\toprule
Metric & Score \\
\midrule
AP & 0.0  \\
AMI & 0.070 \\
ARS & 0.041  \\
FMS & 0.390 \\
Completeness & 0.086 \\
Homogeneity& 0.061\\
V-Measure & 0.071\\
F1 & 0.522 \\
\botrule
\end{tabular}
\end{table}

As with the contingency table of Figure~\ref{figure11}, almost all of the metrics gave scores less than 0.1 to the contingency table of Figure~\ref{figure12}. The FMS metric gave a slightly lower score (0.390) to the contingency table of Figure~\ref{figure12} than it did to the Figure~\ref{figure11} contingency table (0.444). In contrast, the F1 metric gave a slightly higher score (0.522) to the Figure~\ref{figure12} contingency table than it did to the Figure~\ref{figure11} contingency table (0.502).

All of the metrics tested gave similar scores to both contingency tables that had an AP score of zero that were presented here. It is instructive to present two different cases so that the variability in the scores from one contingency table to another can be viewed. 

\subsubsection{Correlation Coefficients} \label{correl coeffs}
\begin{figure}
\centering
\includegraphics[width=0.7\linewidth]{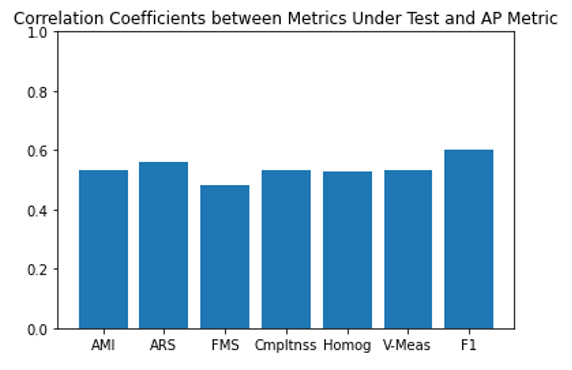}
\caption{Correlation coefficients: low performance, 4x4}
\label{figure13}
\end{figure}

Figure~\ref{figure13} shows a bar plot of the correlation coefficients between each metric under test and the AP metric for the 4x4, low performance contingency tables. 

The correlation coefficients shown in Figure~\ref{figure13} range from 0.483 for FMS to 0.602 for F1. Bobbitt \cite{bobbitt2020} presents a chart, shown in Table~\ref{table9}, that gives verbal descriptions of the strength of relationship for various values of the correlation coefficient.

\begin{table}[h]
\caption{Relationship strength for different correlation coefficients \cite{bobbitt2020}}
\label{table9}
\begin{tabular}{@{}ll@{}}
\toprule
Absolute value of & Strength of relationship \\
correlation coefficient r &  \\
\midrule
$< 0.25$ & No~relationship  \\
$0$~to~$0.5$  & Weak~relationship \\
$0.5$~to~$0.75$  & Moderate~relationship  \\
$r > 0.75$ & Strong~relationship \\
\botrule
\end{tabular}
\end{table}

Based on the values in Table~\ref{table9}, FMS has a weak correlation with the AP metric, and the rest of the metrics have a moderate correlation with the AP metric. 

This section has presented results for contingency tables that represent poor clustering performance. The next three sections present results for contingency tables that represent better clustering performance. The method for generating the higher-performance contingency tables is described earlier in Section~\ref{eval method}. The next three sections present data for different-sized contingency tables—4x4 as was considered in the present section for low-performance contingency tables, as well as 4x6 and 4x2. Each case represents situations in which the clustering algorithm produced different numbers of clusters. 

\subsection{Results: Higher-Performance Case, 4x4 Contingency Tables}
\begin{figure}
\centering
\includegraphics[width=0.8\linewidth]{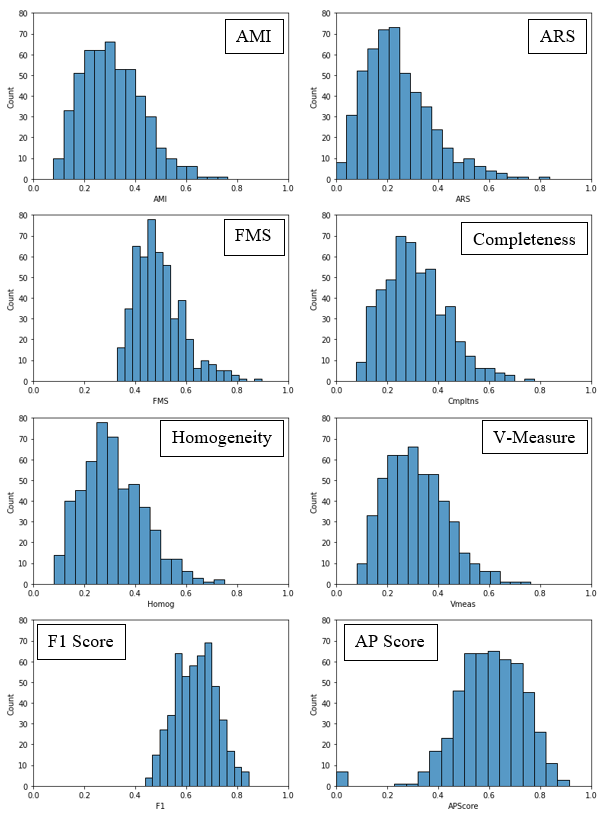}
\caption{Histogram distributions for metrics: higher performance, 4x4}
\label{figure14}
\end{figure}

Figure~\ref{figure14} shows, for each of the metrics, the histogram of scores for the 500 randomly generated 4x4 contingency tables, which represents higher clustering performance. These contingency tables would be produced by clustering algorithms that produce four clusters from data that has four truth classes.

The score distribution plots in Figure~\ref{figure14} are generally shifted toward higher scores as compared with the comparable plots in Figure~\ref{figure8}  for the low-performance 4x4 contingency tables, confirming the designation of higher performance. The peaks’ values in the scores distributions are larger, and the right-hand tails of the distributions, which represent higher scores, are shifted toward higher values. The AP score distribution behaves similarly—there are fewer very low scores, and the overall distribution is shifted toward higher scores.

\begin{figure}
\centering
\includegraphics[width=0.7\linewidth]{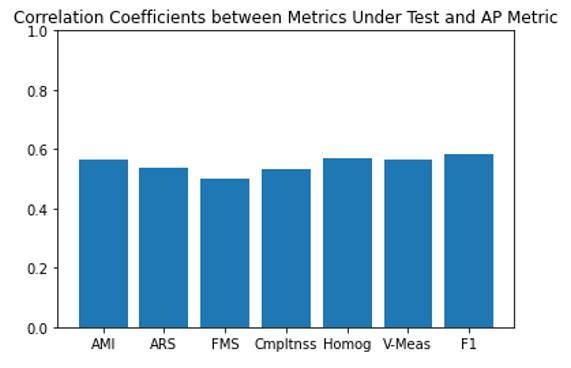}
\caption{Correlation coefficients: higher performance, 4x4 }
\label{figure15}
\end{figure}

Figure~\ref{figure15} shows a bar plot of the correlation coefficients between each metric under test and the AP metric for the 4x4 contingency tables with higher clustering performance. 

The correlation coefficients in Figure~\ref{figure15} are similar to those in Figure~\ref{figure14} for the lower-performance contingency tables. For some metrics, the score increased slightly, and for some it decreased slightly. For all metrics, the correlation coefficient reflects moderate correlation according to the designations in Table~\ref{table9}. 

\subsection{Results: Higher-Performance Case, 4x6 Contingency Tables}
\begin{figure}
\centering
\includegraphics[width=0.8\linewidth]{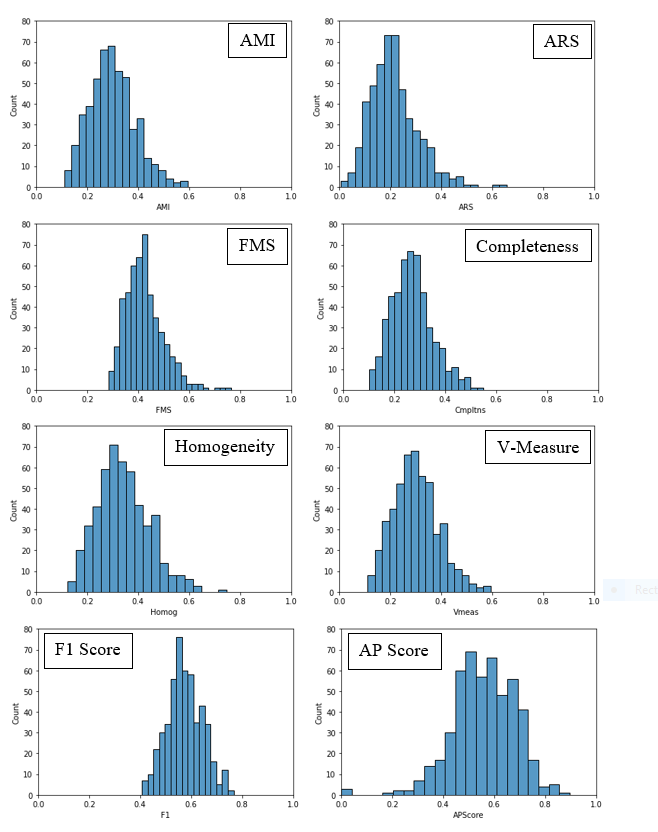}
\caption{Histogram distributions for metrics: higher performance, 4x6}
\label{figure16}
\end{figure}

Figure~\ref{figure16} shows, for each of the metrics, the histogram of scores for the 500 randomly generated 4x6 contingency tables, which represents higher clustering performance. These contingency tables would be produced by clustering algorithms that produce six clusters from data that has four truth classes. Comparing the histogram distributions in Figure~\ref{figure16} with those in Figure~\ref{figure13} for the 4x4 contingency tables shows the effect of having contingency tables with a larger number of clusters.

For all of the metrics, the histogram distributions in Figure~\ref{figure16} are very similar to those in Figure~\ref{figure13}. Thus, increasing the number of clusters produced by clustering algorithms from four to six with four truth classes does not significantly affect the distribution of the scores of any of the metrics tested here.

\begin{figure}
\centering
\includegraphics[width=0.7\linewidth]{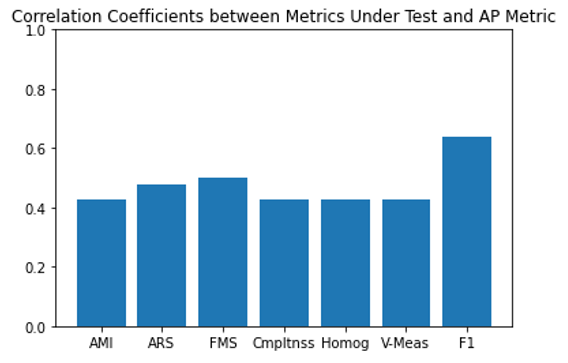}
\caption{Correlation coefficients: higher performance, 4x6}
\label{figure17}
\end{figure}

Figure~\ref{figure17} shows a bar plot of the correlation coefficients between each metric under test and the AP metric for the 4x6 contingency tables with higher clustering performance.

Comparing the bar plot of Figure~\ref{figure17} with the comparable plot in Figure~\ref{figure15} for 4x4 contingency tables shows that for all the metrics except the F1 metric, the correlation with the AP metric decreased when the contingency tables had six clusters instead of four. In contrast, the F1 metric’s correlation increased from 0.583 in the 4x4 case to 0.638 in the 4x6 case. Figure~\ref{figure16} shows that the score distributions of the F1 metric and the AP metric are quite similar in that both have a peak of slightly less than 0.6, and they have a similar distribution of scores larger than 0.6. 

\subsection{Results: Higher-Performance Case, 4x2 Contingency Tables}
\begin{figure}
\centering
\includegraphics[width=0.8\linewidth]{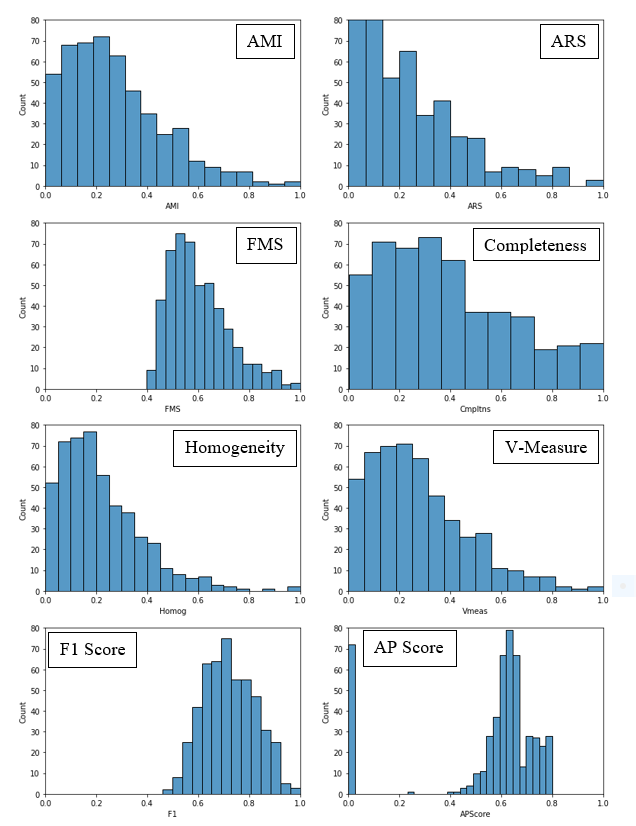}
\caption{Histogram distributions for metrics: higher performance, 4x2}
\label{figure18}
\end{figure}

Figure~\ref{figure18} shows, for each of the metrics, the histogram of scores for the 500 randomly generated 4x2 contingency tables, which represents higher clustering performance. These contingency tables would be produced by clustering algorithms that produce two clusters from data that has four truth classes.

The histograms in Figure~\ref{figure18} for the 4x2 contingency tables differ significantly from the histograms in Figure~\ref{figure14} for the higher-performance 4x4 contingency tables. ARS and the entropy-based metrics, AMI, Completeness, Homogeneity, and V-Measure, had many more scores at the low end of the scale, including many zero scores. That is not surprising, since with 4x2 contingency tables, many of the contingency tables would have the worst-case configuration in which all data samples are in the same cluster, as shown in Figure~\ref{figure7}, for example. ARS and all of the entropy-based metrics except Completeness gave a zero score to that contingency table. Even so, Completeness did give many contingency tables a zero or near-zero score in the 4x2 case. 

Another difference in the behavior of the entropy-based metrics and ARS in the 4x2 case as compared with the 4x4 case is that these metrics had a few more very high scores in the 4x2 case than in the 4x4 case.

The FMS and F1 metrics behaved quite differently than the entropy-based metrics and ARS in the comparison between the 4x4 and 4x2 contingency tables. For these two metrics, the distribution of scores changed little, except that the entire distributions shifted toward higher values in the case of 4x2 contingency tables.

The AP score distribution changed in yet a different way in the 4x2 case as compared with the 4x4 case. In the 4x2 case, it gave very many contingency tables a zero score, due to the prevalence of worst-case contingency matrices, as described earlier. 

Notably, in the 4x2 case, only the AP score gave no scores of 1, whereas all the other metrics did give some scores of 1 or close to 1. This could be explained by the impossibility of getting a high associativity score in contingency matrices with fewer clusters than truth classes. In such cases, it is impossible for there to be a one-to-one matching of clusters with truth classes. This is an important finding, since the presence of fewer clusters than truth classes shows that class differences in the data are being obscured by the clustering algorithm, and the AP scores reflect this fact. 

\begin{figure}
\centering
\includegraphics[width=0.7\linewidth]{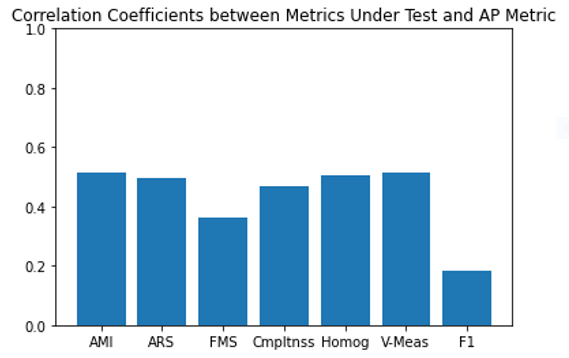}
\caption{Correlation coefficients: higher performance, 4x2}
\label{figure19}
\end{figure}

Figure~\ref{figure19} shows a bar plot of the correlation coefficients between each metric under test and the AP metric for the 4x2 contingency tables with higher clustering performance. 

Comparing Figure~\ref{figure19} for the 4x2 case with Figure~\ref{figure15} for the 4x4 case shows that in the 4x2 case, the correlation coefficient of all of the metrics with the AP score is smaller in the 4x2 case. 

\begin{figure}[!t]
\centering
\includegraphics[width=0.8\linewidth]{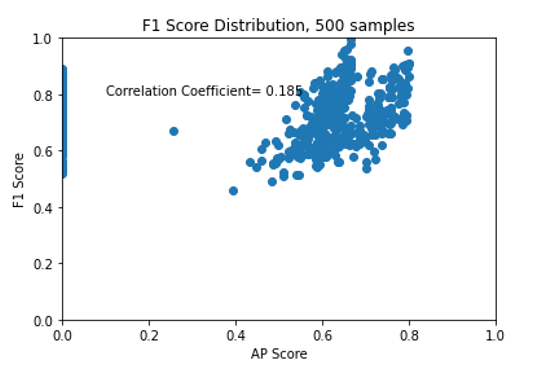}
\caption{Scatter plot for F1 and AP metrics: higher performance, 4x2 }
\label{figure20}
\end{figure}

For the F1 metric, the decline was especially large—from 0.583 in the 4x4 case to 0.185 in the 4x2 case. In the 4x2 case, the F1 metric gave higher scores than the AP score for the large majority of the contingency tables, as shown in the scatter plot of Figure~\ref{figure20}. 

This decreased correlation between the F1 and AP metrics in the 4x2 case is partly due to the many zero scores that the AP metric gave, whereas the F1 metric gave no scores even close to zero in the 4x2 case.

\subsection{Comparison of AP Metric and F1 Metric} \label{compare ap f1}
Of all the metrics under test in this study, only the F1 metric was designed for use with confusion matrices, which are similar in structure to contingency tables. Furthermore, in most test scenarios, it was the most highly correlated with the AP metric. Thus, it is worthwhile to compare these two metrics in detail. 

As mentioned earlier, the metrics in \cite{bib10} are analogous to the contingency table metrics of truth class accuracy (\ref{eqn1}) and cluster purity (\ref{eqn2}). A disadvantage in the truth class accuracy metric becomes apparent when a clustering algorithm produces, for one or more truth values, a few clusters with large populations and several clusters with relatively small populations. In that case, even if the maximum cluster population is large for a particular truth class, the multitude of small cluster populations for that truth class can make the denominator (sum of cluster populations) quite large and thereby reduce the value of truth class accuracy. However, the presence of these multiple low-population clusters would not diminish confidence in the validity of the maximum cluster population value. 

As an example, consider the contingency table of Figure~\ref{figure1}, which contains only the first four columns of the full contingency table, shown in Table~\ref{table10}.

\begin{table}[h]
\caption{Contingency table of Figure~\ref{figure1} with additional columns}
\label{table10}
\begin{tabular}{@{}llllllllll@{}}
\toprule
 Row 1 & 151 & 88 & 72 & 260 & 21 & 0 & 0 & 55 & 13 \\
 Row 2 & 302 & 330 & 0 & 158 & 24 & 0 & 0 & 9 & 0 \\
 Row 3 & 161 & 0 & 313 & 81 & 7 & 0 & 0 & 1 & 65  \\
 Row 4 & 490 & 0 & 101 & 14 & 82 & 0 & 0 & 0 & 4  \\
\midrule
 & C1 & C2 & C3 & C4 & C5 & C6 & C7 & C8 & C9 \\
\botrule
 \end{tabular}
\end{table}

Table~\ref{table10} shows data for nine clouds, each of which is labeled in the bottom row of the table as the letter "C" followed by the cloud number. Clouds 5–9 have much lower populations than clouds 1–4. According to (\ref{eqn12}) for the peakiness metric, the originally calculated values in Table~\ref{table2} would be the same for the contingency table shown in Table~\ref{table10}, since the peakiness metric depends only on the two largest values in a row, which in this case are located in the left-most four columns (i.e., clouds 1–4) for each row. 

However, according to (\ref{eqn1}) for the truth class accuracy metric, this metric is defined as the peak value in a row divided by the sum total of all values in that row, and therefore the value of the truth class accuracy metric would be affected by the presence of the data entries in clouds 5–9. Table~\ref{table11} shows the scores of the peakiness metric and the truth class accuracy for each of the four rows of Table~\ref{table10}.

\begin{table}[h]
\caption{Peakiness and truth class accuracy scores}
\label{table11}
\begin{tabular}{@{}lll@{}}
\toprule
Row & Peakiness & Truth class \\
 &  & accuracy \\
\midrule
1 & 0.419 & 0.394  \\
2 & 0.085 & 0.401  \\
3 & 0.486 & 0.498  \\
4 & 0.794 & 0.709  \\
\botrule
 \end{tabular}
\end{table}

Table~\ref{table11} shows that the largest difference between the scores of peakiness and truth class accuracy is in row 2 of Table~\ref{table10}. For the truth class accuracy metric, the score of 0.401 is similar to the scores of the other three rows for that metric. However, the peakiness metric severely penalizes this row, since the second-largest value, 302, is very close to the second-largest value, 330, and therefore the confidence in 330 as the peak value for that row is low. The truth class accuracy metric does not detect the low confidence in the peak value in row 2. 

The second-largest difference between the scores of peakiness and truth class accuracy is shown in row 4 of Table~\ref{table11}. In this row, the peakiness is quite high due to the presence of one value, 490, which is much larger than all other values in that row. In contrast, the truth class accuracy is lower than the peakiness in this row, due to the presence of several values high population values that increase total population in that row, which increases the denominator of (\ref{eqn1}).

In summary, the peakiness metric explicitly measures the extent to which the peak value in a row is an outlier compared to the rest of the values in that row. The truth class accuracy metric measures the ratio of the peak value in a row to the total population in that row. As a result, the truth class accuracy metric does not detect cases in which the peak value is just slightly larger than other value(s) in that row. Also, truth class accuracy is affected by the presence of several low-population clusters, which are not relevant to the extent to which the peak value is an outlier.

The second component used in the F1 metric  is cluster purity, defined in (\ref{eqn2}). The cluster purity metric is defined very similarly to truth class accuracy in that it seeks to reward high peaks in the truth class populations for each cluster. The combination of the two metrics, truth class accuracy and cluster purity, does reward the presence of cells in the contingency table which are peaks in both the truth class direction and the cluster direction, as illustrated in Figure~\ref{figure2}. However, the AP metric rewards the presence of such cells more explicitly, since (a) the peakiness metric measures the confidence in the peak cluster population values more effectively as described in the previous paragraphs, and (b) the associativity metric more explicitly and directly rewards the presence of cells that are peaks in both the truth class direction and the cluster direction, as can be seen from (\ref{eqn7}), (\ref{eqn8}), and (\ref{eqn9}).

Thus, the peakiness metric is advantageous over the truth class accuracy metric in that it more effectively measures peakiness, and the associativity-peakiness metric pair measures associativity more explicitly and directly than the combination of truth class accuracy metric and cluster purity metric described in Section~\ref{f1 clustering metric}.

\section{Data Analysis}
This section summarizes the findings presented earlier in this paper, organized by test scenario, and discusses how the presented data addresses the objectives of the data presentation, namely, 

\begin{itemize}
\item{What does the AP metric tell us that the scikit-learn and F1 metrics do not? }
\item{How well do the scikit-learn metrics and the F1 metric correlate with the AP metric (i.e., how well do they evaluate associativity)?}
\end{itemize}

Furthermore, the following questions are addressed where applicable:

\begin{itemize}
\item{What about the different metrics accounts for their performance?}
\item{In what situations would the AP metric be most useful and perform better than the scikit-learn metrics?}
\end{itemize}

\subsection{Extreme Cases} \label{extreme cases}
All metrics under test gave a score of 1.0 to the ideal case, and all but three gave a score of 0 to the worst case. As expected, Completeness gave a score of 1 to that worst case, since in that worst case, the clusters from each truth class were entirely contained within a single cluster—the condition for a score of 1 for the Completeness metric. Thus, Completeness alone is not a suitable metric for comparing clustering algorithms, though it could work well as part of the V-Measure metric.

The other two metrics that gave a nonzero score to the worst case were FMS and F1. The definitions of these metrics — (\ref{eqn5}) for FMS, and (\ref{eqn1}) and (\ref{eqn2}) for F1—show that these metrics would not give a zero score to this worst case. In the case of FMS, the score would be zero only when the TP (true positives) is zero, and that would occur only when NO point pairs belong to the same clusters in both the true labels and the predicted labels. In the case of the F1 metric, the score would be zero only when, for all truth classes, the maximum cluster population is zero. This condition would be satisfied only when the entire contingency table consists of zeros. Similarly, the cluster purity would be zero under the same condition. This explains why the F1 metric gave no low scores in this study -- actually no scores below 0.4. Thus, FMS and F1 would not be effective metrics for evaluating algorithms with poor performance.

In the following sections on low- and higher-performance contingency tables, the performance of ARS was similar to that of the entropy-based metrics, and is grouped together with them in the performance analysis. However, the performance of ARS does differ from that of the entropy-based metrics in that its dynamic range was slightly higher than that of the entropy-based metrics in all of the test cases. The histograms for each of the test cases (Figures~\ref{figure8}, \ref{figure14}, \ref{figure16}, and \ref{figure18}) showed that ARS had more instances of very low scores, as well as of very high scores, as compared with the entropy-based metrics, although Figure~\ref{figure18} for the 4x2 contingency table shows that both  ARS  the entropy-based methods had scores ranging all the way from 0 to 1.

\subsection{Low Performance Contingency Tables} \label{low perf cont tables}
ARS, as well as the entropy-based metrics, AMI, Completeness, Homogeneity, and V-Measure, gave low scores to the set of contingency tables—the most common value was about 0.15 for all of them. However, none of these metrics gave a zero score to any of the contingency tables. The metrics not based on entropy, FMS and F1, gave much higher scores—their values were 0.4 and 0.5, respectively. As the analysis from the extreme cases showed, these metrics are not designed to give low scores to contingency tables that indicate low performance. Only the AP metric gave scores of zero, as well as giving more scores exceeding 0.8 than the other metrics did. 

Four selected contingency tables from this test case—two with the highest AP score and two with zero AP score—were selected for more detailed analysis. For the contingency tables with highest AP scores, ARS and the entropy-based metrics all gave very low scores (between 0.2 and 0.3) to one contingency table but gave higher scores (between 0.4 and 0.5) to the other contingency table. This discrepancy, as well as the low correlation scores between these metrics and the AP metric, shows that ARS and the entropy-based methods are measuring something different than associativity and peakiness. The metrics that are not entropy-based, FMS and F1, gave higher scores (ranging from 0.4 to 0.7) for these datasets. However, their scores do not come near the scores that the AP metric gave (0.8) to these very high-performing contingency tables.

Regarding the contingency tables with zero AP scores, the performance of all the metrics was similar to their performance with the worst-case contingency table in Section~\ref{extreme cases}. The non-entropy-based metrics, FMS and F1, did not give low scores to these contingency tables. As mentioned earlier, these two metrics do not regard the occurrence of all truth-class peaks in the same cluster as an undesirable phenomenon. 

During deployment of the clustering algorithm, if one cluster’s total population is much larger than all the rest of the cluster totals, which occurs when the AP score is 0, then the user might think that there is only one cluster and thus only one truth class, and discard the remaining clusters as noise. This would be a severe error, since there are actually multiple truth classes, and thus the case of AP score equal to zero is quite a detrimental situation. Therefore, it is very important for a contingency table metric to identify cases in which the data points are all erroneously grouped into one or two clusters (i.e., many fewer clusters than the number of truth classes).

\subsection{Higher Performance Contingency Tables, 4x6} \label{high perf cont tables 4x6}
Increasing the number of clusters produced by clustering algorithms from 4 to 6 with four truth classes does not significantly affect the distribution of the scores of any of the metrics tested here.

\subsection{Higher Performance Contingency Tables, 4x2}  \label{hi perf cont tables 4x2}
The 4x2 case is similar to the worst case of the Extreme Cases section (Section~\ref{extreme cases}). In the worst case, all data samples are in one cluster. In the 4x2 case, there are only two clusters, so all data samples are members of one of these two clusters. In both of these cases, there is significant obscuration of the class differences among the data samples. As could be expected based on the worst-case performance, ARS and the entropy-based metrics, AMI, Completeness, Homogeneity, and V-Measure, had many more scores at the low end of the scale, including many zero scores. 

For the FMS and F1 metrics, the entire score distributions shifted toward higher values, reaching as high as 1 in this test case. These two metrics give high scores in a test case with only two clusters and four truth classes. This is quite a detrimental performance for the FMS and F1 metrics, since they do not recognize the obscuration of truth classes that results from clustering all the data points into just a few clusters, many clusters fewer than the number of truth classes.

Notably, in the 4x2 case, only the AP score gave no scores of 1, whereas all the other metrics did give some scores of 1 or close to 1. This could be explained by the impossibility of getting a high associativity score in contingency matrices with fewer clusters than truth classes. In such cases, it is impossible for there to be a one-to-one matching of clusters with truth classes. This is an advantage of the AP metric -- it penalizes cluster assignments in which almost all points are in a single cluster, though there are multiple truth classes.

\section{Computational Complexity Analysis}  \label{comput complex}
An important consideration in evaluating clustering metrics is the execution time of the metric, which depends on the computational complexity of the metric. If the metrics are to be calculated using a computationally constrained device, then computational complexity becomes an important consideration. For example, if a reinforcement learning algorithm is running on an edge device that is connected to some environmental sensors, and the value function depends on the clustering performance, then the clustering metric would have to be calculated many times on this edge device. 

Big-O formulas for the computational complexity of the metric are not presented here, since these formulas depend on the particular implementation of the metric. 

For the various metrics presented here, the execution time was measured for each of 500 instances of computation of the metric. For the AMI, ARS, FMS, Completeness, Homogeneity, and V-Measure metrics, the scikit-learn implementation was used. For the F1 score and AP score, the authors’ implementation was used.

The execution time measurement of the scikit-learn implementations included only the time to import the code for the metric and the time to execute the metric. It did not include the time to transform the contingency table to a truth–prediction vector pair.

The computer used to execute the calculations of the metrics is a Dell Precision 7730 laptop. The processor is an Intel Core i7-8850H CPU at 2.60 GHz, with 6 cores and 12 logical processors. The installed physical memory is 32.0 GB. 

\begin{figure}[!t]
\centering
\includegraphics[width=0.8\linewidth]{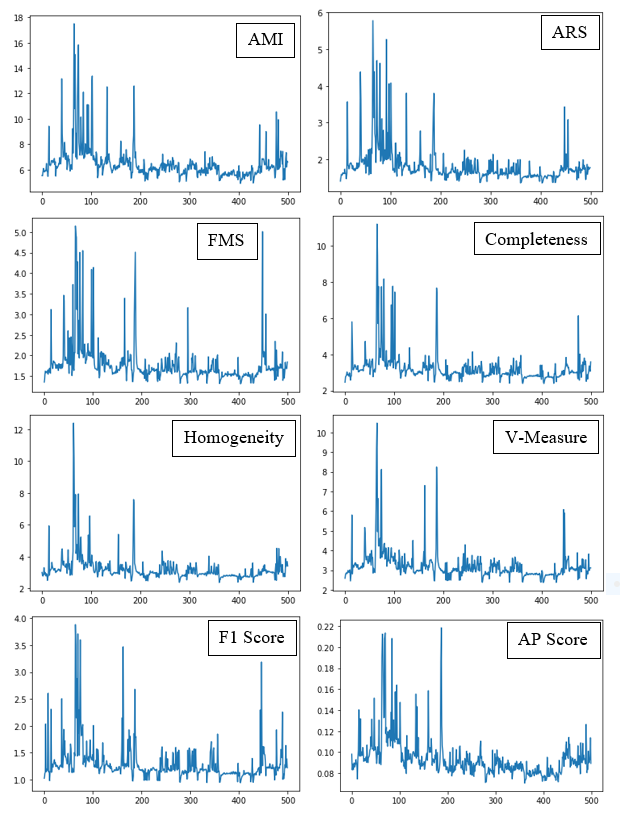}
\caption{Execution time plots; x-axis is instance number; y-axis is time in milliseconds}
\label{figure21}
\end{figure}

The measured computation times in milliseconds for each of the metrics for each of the 500 trials are shown in Figure~\ref{figure21}.

The plots in Figure~\ref{figure21} all show distinct patterns of sharp peaks between instances 0 and 200, and between 400 and 500. A likely explanation for these peaks is that they indicate that the computer which was used to calculate these metrics was running some other processes in the background that slowed the execution of the code used to generate the metrics. A relatively quiescent period occurred between indices 200 and 400. Therefore, the mean values for execution time for each metric is calculated only for values whose index is between 200 and 400. The mean execution times for each metric during this interval are shown in Table~\ref{table12}.

\begin{table}[h]
\caption{Table of mean execution times for the metrics}
\label{table12}
\begin{tabular}{@{}ll@{}}
\toprule
Metric  & Mean execution time (ms)\\
\midrule
AP & 0.085  \\
AMI & 5.967\\
ARS & 1.645  \\
FMS & 1.629 \\
Completeness & 2.985 \\
Homogeneity & 2.967\\
V-Measure & 2.991\\
F1 & 1.197 \\
\botrule
 \end{tabular}
\end{table}

Table~\ref{table12} shows that the entropy-based metrics (AMI, Completeness, Homogeneity, and V-Measure) took the longest, followed by the pairwise-based metrics (ARS and FMS). The fastest metrics were the metrics designed for contingency tables (F1 and AP). Of these two fastest metrics, AP ran in less than 1/10 of the time that it took the F1 metric to run. Thus, for the 200 contingency tables analyzed, the AP metric executed most quickly. Further research to develop analytical expressions for the computational complexity of these metrics would be quite useful, since the data presented in Table~\ref{table12} is for just one value of $N$, the number of samples.

\section{Conclusions}  \label{conclusion}
The equations of the FMS and F1 metrics show that they are both very unlikely to give zero scores or quite low scores, even when evaluating worst-case contingency tables. The lowest score that they gave in this study is about 0.4 on a scale of 0 to 1. These metrics tended to give high scores. The ARS and the entropy-based metrics generally gave lower scores than did FMS, F1, and the AP score.  Thus, ARS and the entropy-based metrics were good at giving low scores to poor performance, and the non-entropy-based metrics were good at giving high scores to good performance, but only the AP metric consistently did both. The AP metric has higher dynamic range, and therefore higher sensitivity, than the scikit-learn metrics and the F1 metric.

The F1 metric was compared with the AP metric. The comparison showed that the AP metric was better at detecting low peakiness in the set of cluster populations for a particular truth class. Also, the AP metric's peakiness score was not reduced by the presence of several low-population clusters for a particular turth value, whereas the F1 metric score was.

The correlation coefficient plots showed that the scikit-learn metrics and the F1 metric had only moderate correlation with the AP metric, generally about 0.5. 

The AP metric’s execution time was over an order of magnitude less than that of all the other metrics tested in the experiment described earlier.

Each metric measures something different. The AP metric's specialization is characterizing performance of clustering algorithms, and predicting their performance when deployed to operate with no class labels. The AP metric has the highest dynamic range and the lowest execution time of all the metrics considered in this paper. These qualities are both very advantageous for a deployed clustering algorithm.

\vskip 0.2in

\bibliography{bibliographyFile}

\end{document}